\pdfoutput=1

\documentclass{article}
\usepackage{arxiv}

\usepackage[utf8]{inputenc} 
\usepackage[T1]{fontenc}    
\usepackage{hyperref}       
\usepackage{url}            
\usepackage{booktabs}       
\usepackage{amsfonts}       
\usepackage{nicefrac}       
\usepackage{microtype}      
\usepackage{lipsum}		
\usepackage{caption} 
\usepackage{graphicx}
\usepackage{natbib}
\usepackage{doi}
\usepackage{amsmath, amssymb, amsfonts}
\usepackage{algorithm}
\usepackage{algpseudocode} 
\usepackage{bbm}           
\usepackage{amsthm}

\newtheorem{theorem}{Theorem}
\newtheorem{definition}{Definition}

\title{Generalisation in Multi-task Offline Q-Learning}

\date{December 14, 2025}

\author{
Manda Kausthubh \\
Department of CSE, IIIT Bangalore\\
Bangalore, KA, India\\
\texttt{manda.kausthubh@iiitb.ac.in}
\And
Raghuram Bharadwaj \\
Department of DSAI, IIIT Bangalore\\
Bangalore, KA, India\\
\texttt{raghuram.bharadwaj@iiitb.ac.in}
}

\renewcommand{\shorttitle}{\textit{arXiv} Template}

\hypersetup{
pdftitle={A template for the arxiv style},
pdfsubject={q-bio.NC, q-bio.QM},
pdfauthor={David S.~Hippocampus, Elias D.~Striatum},
pdfkeywords={First keyword, Second keyword, More},
}

\begin{document}
\maketitle

\begin{abstract}
We study guarantees of offline reinforcement learning in a multitask setting \cite{yu2021conservative}, focusing on Multitask Fitted Q-Iteration (MFQI) \cite{borsa2016learningsharedrepresentationsmultitask} under a shared low-rank structure in action-value functions. Given fixed datasets from multiple tasks over shared environments, we analyze how MFQI jointly learns a common representation and task-specific value functions via Bellman error minimization. Under standard offline RL assumptions, we derive finite-sample guarantees showing that pooling data across tasks improves estimation rates, achieving a $\mathcal{O}(1/\sqrt{nT})$ dependence on the total number of samples. This shows that a simple MFQI algorithm is competitive with similar algorithms in terms of sample complexity. We further consider transfer to a new task that shares the same representation, and show that reusing the learned features reduces the effective complexity of downstream learning. These results clarify when shared representations improve generalization in offline, value-based reinforcement learning. The code has been made available in \texttt{\href{https://github.com/MandaKausthubh/PAC-Analysis-of-MFQI}{[github.com/MandaKausthubh/PAC-Analysis-of-MFQI]}}.
\end{abstract}


\section{Introduction}

In most real-world reinforcement learning (RL) applications, collecting new experience through online interaction is costly, time-consuming, or unsafe. Offline reinforcement learning addresses this challenge by learning decision-making policies from fixed datasets collected by previously deployed behavior policies, without requiring further interaction with the environment. In practice, such datasets often arise from multiple related tasks, motivating the study of \emph{offline multitask reinforcement learning} \cite{yu2021conservative}, where a learner seeks to exploit shared structure across tasks to improve generalization and sample efficiency~\citep{he2025goaloriented, kumar2023offlineqlearningdiversemultitask, ishfaq2024offline, Kumar2022OfflineQO}.

A common assumption \cite{ishfaq2024offline, cheng2022provable, pmlr-v238-oprescu24a} in multitask RL is that tasks share an underlying low-dimensional representation, even though their rewards or dynamics may differ. Learning such shared representations can substantially reduce statistical complexity, particularly when the number of samples per task is limited. Empirically and theoretically, representation sharing has been shown to improve transfer and generalization across tasks \cite{cheng2022provable, ishfaq2024offline, 10.5555/3495724.3497411}, but providing theoretical guarantees in the offline setting remains challenging due to distribution shift, limited coverage, and recursive error propagation through Bellman updates.

Most existing theoretical analyses of multitask representation learning in RL focus on model-based approaches \cite{ishfaq2024offline}, where shared structure is recovered through maximum-likelihood estimation of transition models or latent dynamics. In contrast, many practical algorithms are model-free and value-based \cite{zhu2024on}, relying on fitted Q-iteration and Bellman error minimization. One such model-free algorithm is \emph{Multitask-Fitted Q-iterations} \cite{borsa2016learningsharedrepresentationsmultitask}.

We analyse the offline multitask RL algorithm Multitask Fitted Q-Iterations \cite{borsa2016learningsharedrepresentationsmultitask} from the point of view of sample complexity. In our theoretical analysis, we consider imposing the low-rank realisability assumption (Assumption \ref{assumption1:low-rank}), under which all tasks share a common representation with task-dependent linear decoders. We also provide a short empirical discussion on the non-realisability cases as well. Our theoretical analysis shows us the advantage gained due to the multitask datasets, showing the competitiveness of MFQI\cite{borsa2016learningsharedrepresentationsmultitask} with other similar algorithms \cite{10.5555/3495724.3497411, ishfaq2024offline, cheng2022provable, jin2021pessimism}.


We further consider a downstream offline setting where a new task shares the representation learned from the source tasks. By fixing this representation and learning only the task-specific value function, we obtain a simpler estimation problem that improves sample efficiency compared to learning both components from scratch. Similar downstream offline settings have been studied in previous literature \cite{taylor2009transfer}.

\textbf{Contributions.} Our main contributions are as follows:
\begin{itemize}
    \item We provide a finite-sample analysis of Multitask Fitted Q-iteration (MFQI)\cite{borsa2016learningsharedrepresentationsmultitask} in the offline, model-free setting under a shared low-rank representation assumption.
    \item Our bounds explicitly capture the statistical benefits of multitask data pooling, with estimation error scaling as $O(1/\sqrt{nT})$ under standard realizability and coverage assumptions.
    \item We characterize how Bellman error propagates across the horizon in the multitask setting, highlighting the role of concentrability in offline value-based learning.
    \item We analyze a downstream offline task and show that reusing representations learned from upstream tasks can improve sample efficiency relative to independent learning.
\end{itemize}

\section{Related Work}

\paragraph{Offline Reinforcement Learning.}
Offline or batch reinforcement learning studies policy learning from fixed datasets without further interaction with the environment.
Classical approaches rely on fitted Q-iteration \citep{ernst2005tree, riedmiller2005neural} and subsequent deep extensions such as BCQ, CQL, and IQL \citep{fujimoto2019off, kumar2020conservative, kostrikov2021offline}. 
Theoretical analyses of offline RL have focused on characterizing sample complexity under concentrability assumptions \citep{munos2008finite, antos2008learning, jin2021pessimism}.
Our work extends this line of research to the multitask setting, where shared structure across tasks is exploited to improve statistical efficiency in the offline regime.

\paragraph{Multitask and Representation Learning in RL.}
Representation learning for reinforcement learning has been studied as a means to transfer information across related tasks \citep{bengio2013representation, taylor2009transfer, lazaric2012transfer}.
Recent theoretical progress has formalized multitask representation learning as a low-rank factorization of value or transition functions \citep{agarwal2020optimality, du2019provably}. These works typically assume access to interactive data or known models.
Our formulation builds directly on this multitask structure but focuses on \emph{offline} data settings and on value-based (Q-function) estimation rather than model estimation.

\paragraph{Model-Based Multitask RL.}
Most existing theoretical treatments of multitask offline RL are model-based.
The closest to our work is \citet{ishfaq2024offline, pmlr-v139-domingues21a}, who propose an offline multitask framework that estimates shared low-rank transition models via maximum likelihood estimation (MLE).
Their analysis provides generalization guarantees for multitask model learning but does not extend to model-free Q-learning. 
Our approach replaces the MLE oracle with a Bellman-error minimization oracle, leading to a model-free formulation and analysis that directly connects to practical Q-learning methods.

\paragraph{Model-Free and Value-Based Theoretical Analyses.}
Model-free RL methods, such as Q-learning and actor–critic algorithms, are known to suffer from instability when trained from offline data \citep{levine2020offline}.
Recent theoretical results for single-task offline Q-learning \citep{jin2021pessimism, article} establish finite-sample guarantees via Bellman error concentration.
Our work provides an analysis of the popular method: MFQI  \cite{borsa2016learningsharedrepresentationsmultitask} to the multitask setting and develops new generalization bounds for shared representation learning under mean-squared Bellman error minimization, thereby unifying multitask representation learning and offline value-based RL.

\paragraph{Summary.}
In summary, prior work on multitask reinforcement learning has largely focused on model-based estimation and online interaction, while existing analyses of offline Q-learning are predominantly single-task. Our work complements these directions by studying model-free, value-based learning in the multitask offline setting and providing finite-sample generalization guarantees that parallel those obtained in model-based multitask analyses under comparable assumptions.

\section{Preliminary}

\textbf{Episodic MDP.}
An episodic Markov Decision Process (MDP) is defined as $\mathcal{M} = (\mathcal{S}, \mathcal{A}, H, P, r)$, where $\mathcal{S}$ denotes the state space and $\mathcal{A}$ is a finite action space with cardinality $K = |\mathcal{A}|$. The horizon $H$ specifies the episode length. The transition dynamics are given by a collection of conditional distributions $P = \{P_h\}_{h \in [H]}$, where each $P_h : \mathcal{S} \times \mathcal{A} \to \Delta(\mathcal{S})$. The reward functions are $r = \{r_h\}_{h \in [H]}$ with $r_h : \mathcal{S} \times \mathcal{A} \to [0,1]$. We assume a fixed initial state $s_1$, following standard formulations and notations in the literature \cite{ishfaq2024offline, 10.5555/3495724.3497411}.

A deterministic policy $\pi = \{\pi_h\}_{h \in [H]}$ maps states to actions at each stage, where $\pi_h : \mathcal{S} \to \mathcal{A}$. For a given policy $\pi$, we use $(s_h, a_h) \sim (P,\pi)$ to denote the state–action pair generated at step $h$ by following $\pi$ under dynamics $P$, and expectations are taken with respect to this induced trajectory distribution.

The value function at step $h$ is defined as
$$
V^\pi_{h,P,r}(s) = \mathbb{E}_{(s_h', a_h') \sim (P, \pi)} \left[ \sum_{h'=h}^H r_{h'}(s_{h'}, a_{h'}) | s_h = s \right]
$$
and the corresponding action-value Q-function is
$$
Q^\pi_{h,P,r}(s,a) = \mathbb{E}_{(s_h', a_h') \sim (P, \pi)} \left[ \sum_{h'=h}^H r_{h'}(s_{h'}, a_{h'}) | s_h = s, a_h=a \right]
$$

\paragraph{Multitask Episodic MDPs.}
We consider a collection of $T$ related tasks indexed by $t \in [T]$. Each task is modeled as an episodic MDP $\mathcal{M}^{(t)} = (\mathcal{S}, \mathcal{A}, H, P^{(t)}, r^{(t)})$. The tasks share the same state space, action space, and horizon, but may differ in their transition kernels $P^{(t)}$ and reward functions $r^{(t)}$.

A multitask policy is defined as a collection $\pi = \{\pi^{(t)}\}_{t \in [T]}$, where each $\pi^{(t)} = \{\pi_h^{(t)}\}_{h \in [H]}$ governs decisions for task $t$. The corresponding task-specific value functions are
$$
V^{\pi,t}_{h,P,r}(s) = \mathbb{E}_{(s_h', a_h') \sim (P, \pi^{(t)})} \left[ \sum_{h'=h}^H r^{(t)}_{h'}(s_{h'}, a_{h'}) | s_h = s \right]
$$
Similarly, the task-specific Q-functions are
$$
Q^{\pi,t}_{h,P,r}(s,a) = \mathbb{E}_{(s_h', a_h') \sim (P, \pi^{(t)})} \left[ \sum_{h'=h}^H r^{(t)}_{h'}(s_{h'}, a_{h'}) | s_h = s, a_h=a \right]
$$

\subsection{Multitask Offline Q-Learning with Shared Representations}

In multitask offline Q-learning, we are given fixed datasets collected from $T$ related tasks, where each task corresponds to an MDP that shares a common state–action space but may differ in rewards and/or dynamics. The goal is to exploit shared structure across tasks to improve sample efficiency and generalization. This framing is similar to other general offline RL framing that exists in the literature \cite{levine2020offline, fujimoto2019off, kumar2020conservative, sutton2014introduction}.

We assume that the optimal action-value functions admit a shared representation: there exists a feature encoder $\phi : \mathcal{S} \times \mathcal{A} \rightarrow \mathbb{R}^d$ such that for each task $t$,
\begin{equation}
Q^{*,t}(s,a) = \langle \phi(s,a), w^{*,t} \rangle,
\end{equation}
where $w^{*,t}$ are task-specific parameters.

Learning proceeds via a multitask extension of Fitted Q-Iteration (FQI) \cite{ernst2005tree}, called Multitask Fitted Q-Iterations (MFQI) \cite{borsa2016learningsharedrepresentationsmultitask}. At each iteration $k$, for every transition $(s,a,r,s')$ in the dataset for task $t$, we construct bootstrapped targets using the current Q-function estimate:
\begin{equation}
y^{(k)}_t(s,a) = r_t(s,a) + \gamma \max_{a'} Q^{(k)}_t(s', a').
\end{equation}

Given these targets, the update step reduces to a supervised multitask regression problem. Specifically, we jointly learn the shared representation $\phi$ and task-specific parameters $\{w_t\}_{t=1}^T$ by minimizing the empirical squared error across all tasks:
\begin{equation}
\label{equation3:MTMSELoss}
\frac{1}{nT} \sum_{t=1}^T \sum_{i=1}^n 
\left(
\langle \phi(s^{(i,t)}, a^{(i,t)}), w_t \rangle - y^{(k)}_t(s^{(i,t)}, a^{(i,t)})
\right)^2.
\end{equation}

While the original MFQI \cite{borsa2016learningsharedrepresentationsmultitask} doesn't explicitly state any loss function and creates a generic framework, we consider the Multitask loss to be as stated in \ref{equation3:MTMSELoss} as it is widely used and as it provides a well-defined empirical risk minimization problem. This choice aligns with standard fitted Q-iteration and facilitates theoretical analysis through squared Bellman error minimization. This objective corresponds to minimizing the empirical Bellman residual at each iteration, while enforcing a shared representation across tasks. Crucially, the learning procedure alternates between (i) computing Bellman targets using the current Q-functions and (ii) solving a structured multitask regression problem, rather than directly optimizing a single fixed Bellman error objective.

The shared encoder $\phi$ couples learning across tasks, enabling statistical gains through data pooling and improving generalization in the low-sample regime.

\subsection{Role of the Concentrability Coefficient}
\label{subsec:lambda_max}

A central difficulty in offline reinforcement learning is distribution shift between the data-generating distribution and the distributions induced by Bellman backups.
Since learning is performed using samples from a fixed behavior policy, errors measured under the behavior distribution may be amplified when propagated backward through the Bellman operator. To tackle this, we utilise the popular concentrability coefficient \cite{10.5555/3041838.3041909, munos2008finite}.

Let $\mu_b$ denote the state--action distribution induced by the behavior policy.
For any policy $\pi$ and stage $h$, let $\mu_h^\pi$ denote the state--action distribution induced at stage $h$ by following $\pi$.
We quantify distribution shift using a concentrability coefficient
\[
\lambda_{\max}
\;\triangleq\;
\sup_{h \in [H]} \sup_{\pi}
\left\|
\frac{\mu_h^\pi}{\mu_b}
\right\|_\infty,
\]
which upper bounds the density ratio between Bellman-induced distributions and the offline data distribution. We assume a coverage condition such that the behavior distribution dominates all policy-induced distributions.

In the online or on-policy setting, Bellman operators are contractive in suitable norms, leading to controlled error propagation.
In contrast, in the offline setting, contraction may fail under $\mu_b$, and $\lambda_{\max}$ appears as a multiplicative factor in generalization bounds.
Consequently, the bounds in Theorems~\ref{theorem1:finite_class_generalisation} and~\ref{theorem2:downstream_generalisation} scale polynomially with $\lambda_{\max}$, reflecting the inherent difficulty of offline value-based learning under distribution shift.
When the behavior policy provides sufficient coverage, $\lambda_{\max}$ remains bounded, yielding meaningful finite-sample guarantees.

This multitask empirical-risk objective performs joint representation learning and value-function regression across all tasks, providing a shared encoder that generalizes to unseen downstream environments. To ensure bounded sample complexity, we adopt the \emph{coverage condition} standard in offline RL, assuming that each behavior policy $\pi_b^{(t)}$ sufficiently covers the state--action distribution induced by optimal or target policies. Under this assumption, our analysis establishes finite-sample generalization guarantees for the learned representation $\hat\phi$ and task-specific Q-functions $\hat Q^{(t)}$, showing that shared representations learned via multitask Q-learning achieve comparable sample efficiency to multitask model-based approaches. In Section~\ref{sec:theory}, we formalize this result and provide our main finite-sample bound, demonstrating that the proposed multitask Q-learning framework admits strong theoretical guarantees in the offline setting. Similar methods have been used to derive bounds on offline RL in the literature \cite{Chen2019InformationTheoreticCI}.

\begin{algorithm}[t]
\caption{Multi-task Fitted Q-Iteration \cite{borsa2016learningsharedrepresentationsmultitask}}
\label{algo:MTFQI}
\begin{algorithmic}[1]

\Require $\mathcal{D} = \bigcup_{t=1}^T \mathcal{D}_t \sim (\mu, \mathcal{P})$
\Statex \hfill

\State Initialize parameters $\Theta \gets \Theta_0$, iteration counter $k \gets 0$

\For{$h = H$ \textbf{down to} $1$}
    \Repeat
        \State \textbf{Compute targets:}
        \For{$t = 1$ \textbf{to} $T$}
            \State
            $\mathcal{Y}_t^{(k+1)} \gets 
            \left\{
            R_t(s,a) + \gamma \max_{a'} Q^{(k)}(s',a')
            \;\middle|\; (s,a,s') \in \mathcal{D}_t
            \right\}$
        \EndFor

        \State \textbf{Multi-task update:}
        \State
        $\Theta^{(k+1)} \gets 
        \operatorname{MTL}\!\left(
        \bigcup_{t=1}^T \mathcal{D}_t,\;
        \bigcup_{t=1}^T \mathcal{Y}_t^{(k)}
        \right)$

        \State $d\Theta \gets \lVert \Theta^{(k+1)} - \Theta^{(k)} \rVert$
        \State $k \gets k + 1$
    \Until{$d\Theta < \epsilon$ \textbf{or} $k \ge \text{MaxIter}$}
\EndFor

\State \Return $\Theta = \{\theta_t\}_{t=1}^T$, where $Q_t(s,a) = f_{\theta_t}(s,a)$

\end{algorithmic}
\end{algorithm}

\section{Upstream Learning in MFQI}


In this section, we review the offline multitask Q-learning algorithm presented in MFQI \cite{borsa2016learningsharedrepresentationsmultitask}. This work builds upon, focusing on its formulation in the low-rank setting, and summarizes the theoretical properties relevant to our analysis.

\subsection{Multitask Fitted Q-Iteration Algorithm}
\label{sec:algorithm_design}

The details of the algorithm MQRL (Multitask Q-Learning for Representation Learning) \cite{borsa2016learningsharedrepresentationsmultitask} are depicted in Algorithm~\ref{algo:MTFQI}. The agent receives all offline datasets $\{\mathcal{D}^{(t)}\}_{t \in [T]}$ and jointly estimates the shared encoder $\phi_h$ and the task-specific decoders $\{\psi_h^{(t)}\}_{t \in [T]}$ by minimizing the empirical Bellman error across all tasks at each stage $h$:
\begin{align}
\label{eq:mqrl_oracle}
(\hat{\phi}_h, \hat{\psi}_h^{(1)}, \dots, \hat{\psi}_h^{(T)})=\arg\min_{\phi_h \in \Phi, \psi_h^{(1)}, \dots, \psi_h^{(T)} \in \Psi} \hat{\mathcal{L}}(\phi_h, \{\psi_t\}_{t\in[T]})
\end{align}
This objective acts as a \emph{multitask empirical-risk minimization oracle}, serving as the offline Q-learning analogue of the maximum-likelihood oracle used in prior multitask model-based work \cite{10.5555/3495724.3497411, pmlr-v139-domingues21a}. In practice, the optimization in \ref{eq:mqrl_oracle} can be efficiently approximated whenever the function classes $\Phi$ and $\Psi$ are parameterized by neural networks or other differentiable architectures, allowing gradient-based minimization. The shared encoder $\phi_h$ aggregates statistical information across all tasks, thereby promoting representation sharing, while each decoder $\psi_h^{(t)}$ specializes to its respective task to approximate its local value function. In Algorithm \ref{algo:MTFQI} we prefer $\Theta$ to represent the combination of $\phi$ and $\omega_h^{(t)}$.

\subsection{Theoretical Results}
\label{sec:theory}
The notations, ideas in this section are inspired by \cite{borsa2016learningsharedrepresentationsmultitask, ishfaq2024offline}. To facilitate the analysis, we adopt a realizability assumption \cite{ishfaq2024offline, 10.5555/3495724.3497411} that is standard in the low-rank Q-learning literature. Similar variations of the low-rank assumptions can be found in the literature \cite{10.5555/3495724.3497411, pmlr-v238-oprescu24a}.

\textbf{Assumption 3.1 (Low-rank realizability) \cite{10.5555/3495724.3497411, ishfaq2024offline}.} \label{assumption1:low-rank}
There exists a shared representation $\phi^* \in \Phi$ and task- and stage-specific weight vectors $w_h^{(*,t)} \in \Psi$ such that, for all $h \in [H]$ and $t \in [T]$,
\[
Q_h^{*,t}(s,a) = \langle \phi^*(s,a), w_h^{(*,t)} \rangle.
\]
We assume $\|\phi(s,a)\|_2 \le 1$ for all $\phi \in \Phi$ and $\|w\|_2 \le 1$ for all $w \in \Psi$.

For simplicity, assume that $\Phi$ and $\Psi$ are finite sets. The main theorem here is the following:

\begin{theorem}
\label{theorem1:finite_class_generalisation}
Under the low-rank MDP structure (Assumption 3.1), we establish finite-sample generalization bounds for multitask offline Q-learning \cite{borsa2016learningsharedrepresentationsmultitask}.

\begin{itemize}
    \item[\textit{(a)}] Under realizability, with probability at least \(1-\delta\), for any \(h \in [H]\),
    $$
    \frac{1}{T} \sum_{t=1}^T \mathbb{E}_{(s_h,a_h) \sim \mu_b^{(t)}} \left[ \|\hat{Q}_h^t - Q^{\pi_b^t}_h\|^2 \right] \leq B \sqrt{\frac{2\log(2|\Phi||\Psi|^T H / \delta)}{nT}}.
    $$

    \item[\textit{(b)}] With probability \(1-\delta\), for any \(h \in [H-1]\),
    \begin{align}
        ||\hat Q_h - Q_h^*||_{L_2(\mu_b)} &\leq \sqrt{2\lambda_{\max}}||\hat Q_{h+1} - Q^*_{h+1}||_{L_2(\mu_b)} + \sqrt{\epsilon_{\text{irred}}(|\Phi||\Psi|^T)}\nonumber\\ 
        &\quad\quad+\ \sqrt{\mathcal{O}\left( \frac{2B}{3}\log\left( \frac{2|\Phi||\Psi|^T}{\delta} \right)+ \sqrt{\frac{4B^2}{9}\log^2\left( \frac{2|\Phi||\Psi|^T}{\delta} \right) + 8\sigma^2\log\left( \frac{2|\Phi||\Psi|^T}{\delta} \right)} \right)} \nonumber
    \end{align}

    \item[\textit{(c)}] Averaging across tasks, with high probability,
    $$
    \frac{1}{T} \sum_{t=1}^T \mathbb{E}_{\mu_b^{(t)}} \left[ \|\hat{Q}_1^t - Q_1^{*,t}\|_{L_2(\mu_b)} \right] \lesssim \tilde{O}\left( H \mathcal{E}_{\text{Approx}} + H^2 \lambda_{\max,\sup} \sqrt{\frac{|\Phi||\Psi|^T}{nT}} + H^3 \lambda_{\max,\sup} \frac{|\Phi||\Psi|^T}{nT} \right),
    $$
    where \(\mathcal{E}_{\text{Approx}}\) bounds the layer-wise approximation error, and \(\lambda_{\max,\sup}\) is the supremum concentrability coefficient handling distribution shift. This final bound unrolls the recursion in (b), yielding polynomial dependence on the horizon $H$.
\end{itemize}
\end{theorem}


\paragraph{Remarks.}
Part~\textit{(a)} provides a uniform generalization bound for the fitted Q-function at a fixed stage, highlighting the $1/\sqrt{nT}$ improvement obtained by pooling data across tasks.
Part~\textit{(b)} characterizes how estimation error propagates backward through Bellman updates in the offline setting, with the concentrability coefficient $\lambda_{\max}$ accounting for distribution shift. This also 
Part~\textit{(c)} unrolls this recursion across the horizon to obtain a bound on the initial-stage error, with polynomial dependence on $H$ and explicit dependence on the multitask sample size.
Complete proofs are deferred to the appendix.

\subsection{Generalization Bounds for Infinite Hypothesis Sets}

When the hypothesis spaces for the encoder $\Phi$ and the decoders $\Psi$ are infinite, such as those parameterized by continuous function approximators, the Union Bound approach utilized in Theorem \ref{theorem1:finite_class_generalisation} is no longer applicable. We must instead rely on uniform convergence theory based on the Rademacher complexity \cite{bartlett2002rademacher, ShalevShwartzBenDavid2014, Koltchinskii2002EmpiricalMD} to quantify the capacity of the Multitask Hypothesis Space $\mathcal{F} = \Phi \times \Psi^T$.

\subsubsection{Multitask Loss Class and Rademacher Complexity}

The generalization gap is determined by the complexity of the class of functions that measure the squared Bellman residual. We define the multitask loss function class, $\mathcal{G}$, as:
$$
\mathcal{G} = \left\{ g_{f} : f \in \mathcal{F} \right\}
$$
where $g_{f}$ is the squared Bellman residual for a function $f = (\phi, \{\psi^{(t)}\}) \in \Phi \times \Psi^T$ at step $h$:
$$
g_{f}(s, a, y) = \frac{1}{T}\sum_{t=1}^{T} \left( Q_{h}^{t}(s,a; f) - \text{y} \right)^2
$$
Here, $y = T_{h}^{*} \hat{Q}_{h+1}^{t}$ is the empirical Bellman target used for the regression. I would like the reader to refer to the appendix section for a detailed explanation of Raedmacher complexity and the associated analysis and proof.

\begin{definition}[Expected Rademacher Complexity]
The Expected Rademacher Complexity of the loss class $\mathcal{G}$, denoted $\mathcal{R}(\mathcal{G})$, measures the ability of $\mathcal{G}$ to fit random noise on the training data:
$$
\mathcal{R}(\mathcal{G}) = \mathbb{E}_{\mathcal{D}, \mathbf{\sigma}} \left[ \sup_{g \in \mathcal{G}} \frac{1}{nT} \sum_{t=1}^{T} \sum_{i=1}^{n} \sigma_{i}^{(t)} g(s_h^{(i,t)}, a_h^{(i,t)}, y_h^{(i,t)}) \right]
$$
where $\mathbf{\sigma} = \{\sigma_i^{(t)}\}_{i=1, t=1}^{n, T}$ are independent Rademacher random variables ($\text{Pr}(\sigma=\pm 1)=1/2$).
\end{definition}

\subsubsection{Generalization Bound via Rademacher Complexity}

By applying the standard generalization theorem (e.g., using McDiarmid’s inequality \cite{Mohri2018}, and the Rademacher complexity bound \cite{bartlett2002rademacher, mohri2018foundations}), we can replace the cardinality-dependent term in Theorem \ref{theorem1:finite_class_generalisation} (c) with the complexity measure $\mathcal{R}(\mathcal{G})$.

We define the (average) suboptimality gap at stage $h$ as
\[
\Delta_h \;\triangleq\; \frac{1}{T} \sum_{t=1}^T 
\mathbb{E}_{(s_h,a_h)\sim \mu_b^{(t)}}
\!\left[\,
\|\hat Q_h^t(s_h,a_h) - Q_h^{*,t}(s_h,a_h)\|_{L_2(\mu_b)}
\,\right].
\]
For the downstream task, $\Delta_h^{\text{downstream}}$ denotes the corresponding quantity for the single novel task.


\begin{theorem}
\label{theorem2:downstream_generalisation}
Consider a downstream offline task that shares the same low-rank representation as the upstream multitask setting.
Let $\hat\phi$ denote the representation learned from the upstream tasks, and let $\mathcal{G}$ be the corresponding class of task-specific decoders.
Then, with probability at least $1-\delta$, the downstream suboptimality gap at the initial stage satisfies
\[
\Delta_1^{\text{downstream}}
\;\lesssim\;
H \lambda_{\max,\sup} \epsilon_{\text{irred}}^{\text{eff}}
\;+\;
H^2 \lambda_{\max,\sup} \mathcal{R}(\mathcal{G})
\;+\;
H^3 \lambda_{\max,\sup} \frac{\log(1/\delta)}{n},
\]
where $\epsilon_{\text{irred}}^{\text{eff}}$ denotes the effective irreducible error induced by using the upstream representation $\hat\phi$,
$\mathcal{R}(\mathcal{G})$ is the Rademacher complexity of the decoder class $\mathcal{G}$,
and $\lambda_{\max,\sup}$ is the supremum concentrability coefficient.
\end{theorem}

This result shows that, when the representation is fixed using upstream multitask data, downstream learning reduces to estimating a lower-complexity decoder class. In particular, the statistical term depends on the Rademacher complexity $\mathcal{R}(\mathcal{G})$, rather than the complexity of the full hypothesis class used during upstream learning. For linear decoders over bounded representations, $\mathcal{R}(\mathcal{G}) = O(1/\sqrt{n})$, yielding improved statistical rates compared to learning the representation from scratch.

\subsection{Bounds under Non-Realisability scenarios}
Our theoretical analysis assumes realizability of the underlying MDP within the chosen function class. When this assumption \ref{assumption1:low-rank} is violated, our guarantees for \ref{algo:MTFQI} extend in a standard manner by incurring an additional approximation error term. In particular, if the true transition dynamics admit an $\epsilon-$ approximation within the model class, then the resulting policy suboptimality increases by at most $\mathcal{O}(\mathrm{poly}(H) \epsilon_{\text{Approx}})$ \cite{munos2008finite, antos2008learning}. Empirically, we observe that the proposed method remains stable under moderate misspecification, suggesting robustness beyond the realizable regime.

\section{Experiments}

The goal of our experiments is not to demonstrate empirical performance, but to provide controlled sanity checks that illustrate the scaling behavior predicted by the theoretical results.
In particular, we focus on how estimation error varies as a function of the number of tasks $T$, the number of samples per task $n$, and the horizon $H$, while holding other parameters fixed.

\paragraph{Experimental setup.}
We consider synthetic multitask Markov decision processes constructed to satisfy the low-rank realizability assumptions used in our analysis.
Each task shares a common representation of dimension $d$, with task-specific linear decoders.
Offline datasets are generated using fixed behavior policies, and learning is performed using multitask fitted Q-iteration with squared Bellman error minimization.
Unless otherwise stated, results are averaged over multiple independent runs to reduce stochastic variability.

\paragraph{Scaling with the number of tasks.}
We first study how the estimation error varies as the number of tasks $T$ increases while keeping the number of samples per task $n$ and the horizon $H$ fixed.
Consistent with Theorem~\ref{theorem1:finite_class_generalisation}, the observed error decreases as $T$ grows, exhibiting the expected $O(1/\sqrt{nT})$ dependence due to multitask data pooling.

\paragraph{Scaling with the number of samples per task.}
Next, we vary the number of samples per task $n$ while fixing $T$ and $H$.
As predicted by the theory, the estimation error decreases at a rate proportional to $O(1/\sqrt{n})$, reflecting standard statistical scaling in value-based learning.

\paragraph{Scaling with the horizon.}
Finally, we examine the dependence of the estimation error on the horizon $H$.
We observe polynomial growth in $H$, consistent with the horizon-dependent terms appearing in our theoretical bounds.
This behavior highlights the effect of recursive error propagation in offline value-based learning.

We emphasize that these experiments are conducted in simplified settings where the modeling assumptions hold by construction, and are intended solely to illustrate qualitative scaling trends rather than to serve as comprehensive empirical evaluations.

\begin{figure}[t]
\centering

\begin{minipage}[t]{0.33\textwidth}
    \centering
    \includegraphics[width=\textwidth]{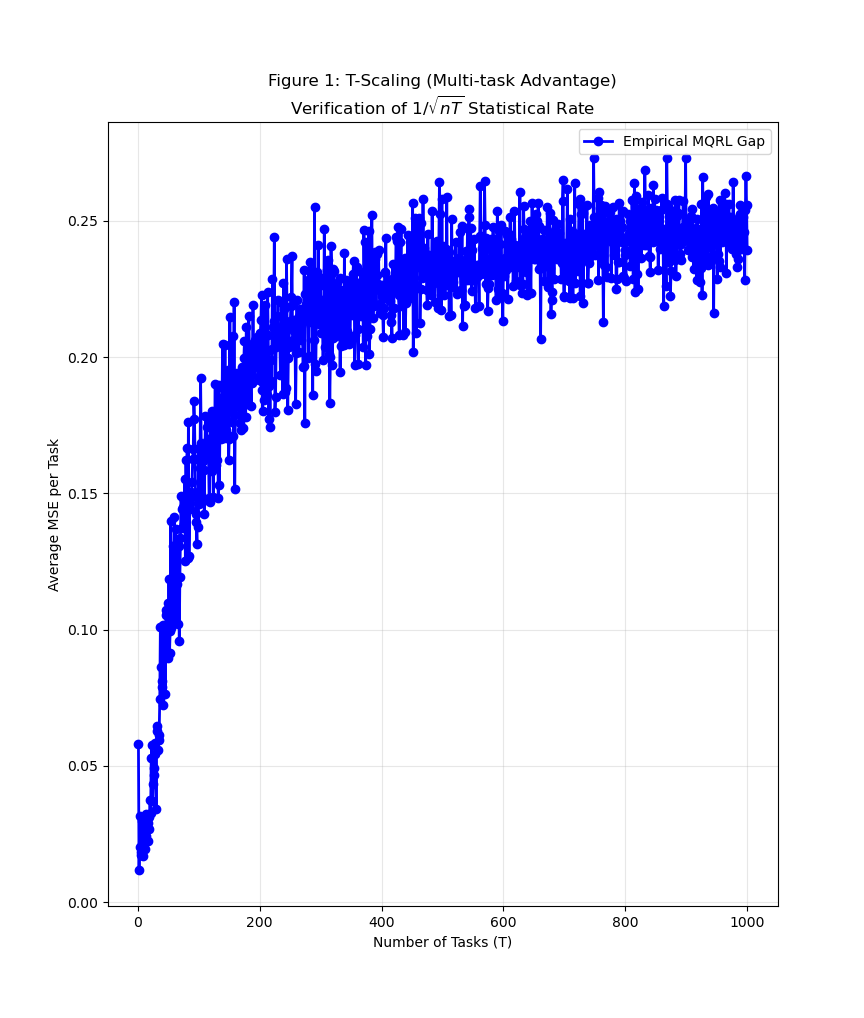}
\end{minipage}
\hfill
\begin{minipage}[t]{0.33\textwidth}
    \centering
    \includegraphics[width=\textwidth]{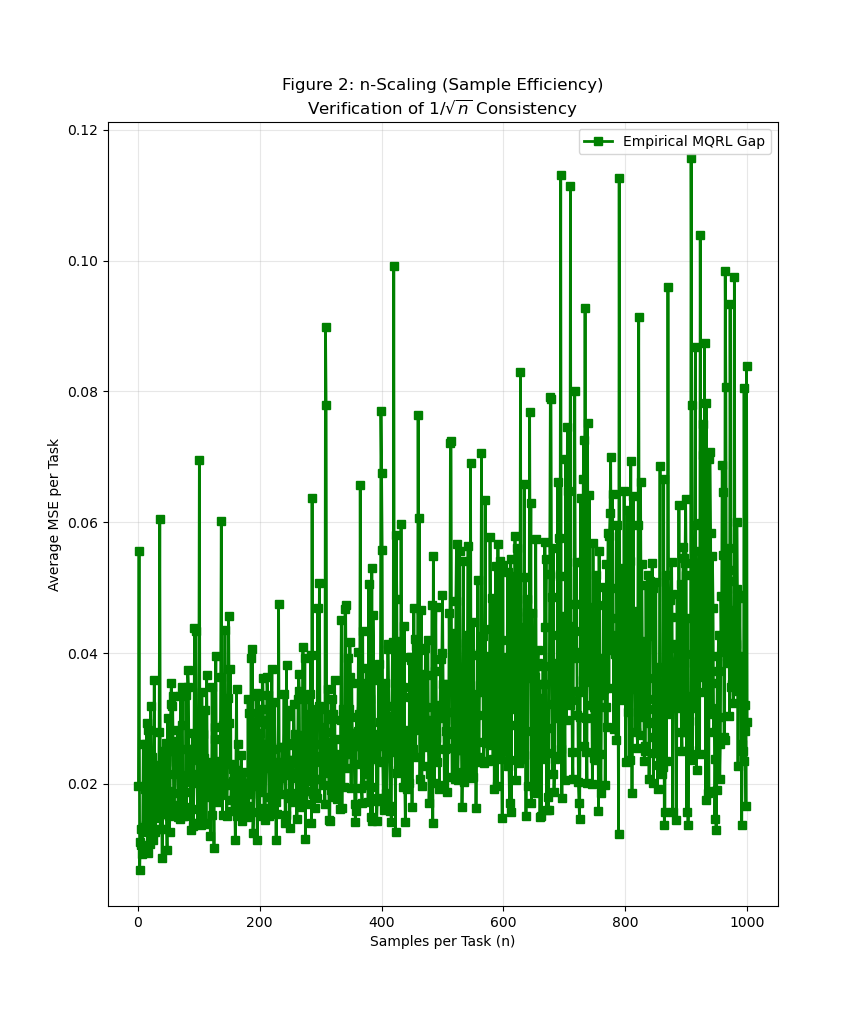}
\end{minipage}
\hfill
\begin{minipage}[t]{0.33\textwidth}
    \centering
    \includegraphics[width=\textwidth]{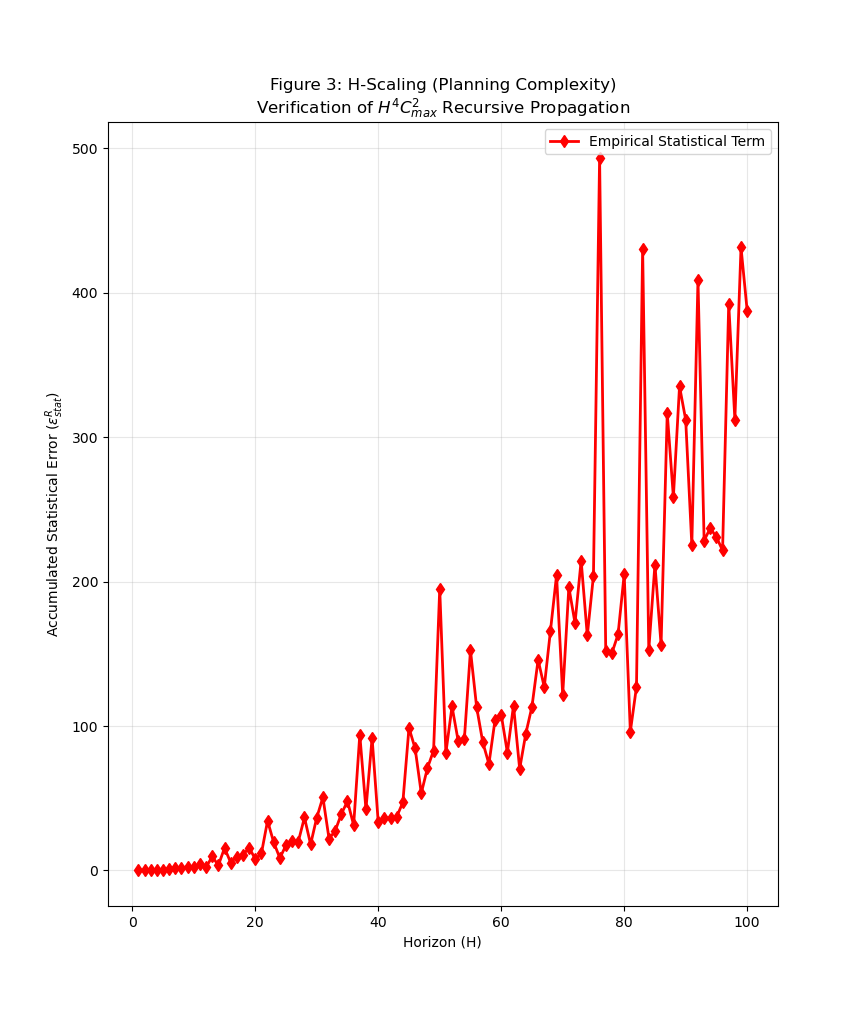}
\end{minipage}

\caption{
Scaling behavior of the proposed method. 
\textbf{Left:} Net Multitask efficiency $(nT\times MSE)$ with increasing $T$ (fixed $n=200, H=5$). 
\textbf{Middle:} Sample consistency with increasing $n$ (fixed $T=5, H=5$). 
\textbf{Right:} Polynomial horizon dependence with increasing $H$ (fixed $T=5, n=500$).
}
\label{fig:scaling-laws}
\end{figure}

When varying the number of tasks $T$ while fixing the number of samples per task $n$ and the horizon $H$, we observe a clear decrease in estimation error as $T$ increases.
Since the total amount of offline data grows linearly with $T$ (total samples $= nT$), Theorem~1(b) \ref{theorem1:finite_class_generalisation} predicts a statistical term scaling as $O(1/\sqrt{nT})$, which reduces to $O(1/\sqrt{T})$ in this regime.
After unrolling the recursion in Theorem~1(c), the dominant contribution to the suboptimality gap $\Delta_1$ scales as $O(H^2 \lambda_{\max}/\sqrt{nT})$.
The observed downward trend in error as $T$ grows is consistent with this predicted behavior and illustrates the benefit of pooling data across tasks when learning a shared representation.

Importantly, the error decreases with increasing $T$, rather than increasing, which would indicate negative transfer or model misspecification.
The observed fluctuations are expected due to stochastic optimization and finite-sample effects, but they do not obscure the overall scaling trend.

Next, we fix the number of tasks $T=5$ and the horizon $H=5$, and vary the number of samples per task $n$.
In this setting, the total dataset size grows linearly with $n$, and Theorems~1(b) and~1(c) predict a leading statistical term scaling as $O(1/\sqrt{n})$.
Consistent with this prediction, the estimation error decreases as $n$ increases, despite noticeable variance across runs.
These fluctuations are expected in multitask training with finite $T$ and do not alter the qualitative $1/\sqrt{n}$ dependence.

Finally, when varying the horizon $H$ while holding $T$ and $n$ fixed, we observe polynomial growth in the estimation error.
This behavior is consistent with the horizon-dependent terms in Theorem~1(c) \ref{theorem1:finite_class_generalisation}, and rules out exponential error propagation.
Overall, these experiments illustrate that multitask offline Q-learning exhibits the qualitative scaling behavior predicted by the theory with respect to the number of tasks, samples per task, and planning horizon.

\section{Conclusion}


In this work, we examined multitask offline reinforcement learning in settings where tasks share a common low-rank structure in their action-value functions. We focused on a model-free, value-based approach based on Multitask Fitted Q-Iteration (MFQI)~\cite{borsa2016learningsharedrepresentationsmultitask}, and described how it can be formulated to jointly learn a shared representation along with task-specific value functions from offline data. Under standard realizability and coverage assumptions, we discussed the corresponding finite-sample guarantees and how they capture the effect of pooling data across tasks. In particular, these results illustrate how shared representations can improve statistical efficiency in the multitask offline setting. Our presentation places MFQI in the context of existing theoretical work and highlights its connection to prior analyses of multitask representation learning in reinforcement learning~\cite{cheng2022provable, ishfaq2024offline, 10.5555/3495724.3497411}.

A key takeaway from our analysis is that the proposed MFQI-based approach attains comparable suboptimality bounds, in order, to existing theoretical results for multitask offline reinforcement learning \cite{10.5555/3495724.3497411, ishfaq2024offline} under related structural assumptions \ref{assumption1:low-rank} \cite{pmlr-v238-oprescu24a, ishfaq2024offline, pmlr-v139-domingues21a}. In particular, under the low-rank realizability \ref{assumption1:low-rank} and coverage conditions considered here, our bounds exhibit polynomial dependence on the horizon and scale as $O(1/\sqrt{nT})$ \ref{theorem1:finite_class_generalisation} with the total number of samples, reflecting the effect of aggregating data across tasks.

Importantly, these guarantees are established in a model-free, value-based setting with shared representation learning, and therefore complement prior analyses that focus on either single-task offline Q-learning or model-based multitask methods. Our results show that, within this formulation, exploiting shared structure across tasks can improve statistical efficiency without worsening the dominant scaling in the sample size or horizon. Theorems: \ref{theorem1:finite_class_generalisation}, \ref{theorem2:downstream_generalisation}

We further analyzed a downstream offline learning setting in which a novel task shares the representation learned from upstream multitask data. In this case, fixing the learned representation reduces the effective complexity of the hypothesis class for the downstream problem, leading to improved sample efficiency relative to learning both representation and value function from scratch, while retaining the same order of horizon dependence. Overall, our findings demonstrate that Multitask Fitted Q-iterations (MFQI) \cite{borsa2016learningsharedrepresentationsmultitask} via value-based methods admit controlled finite-sample guarantees under standard assumptions.

\begin{table}[t]
\centering
\caption{Comparison of final suboptimality/estimation error bounds for offline and multitask reinforcement learning methods. Here $T$ denotes the number of tasks, $n$ the number of samples per task, $H$ the horizon, and $\lambda$ a concentrability coefficient. Logarithmic factors are omitted.}
\label{tab:comparison}
\begin{tabular}{lcc}
\toprule
Method & Setting & Final Error Bound \\
\midrule
Our bound (MFQI) 
& Offline, multitask 
& $\tilde{O}\!\left({H^2 \lambda}/{\sqrt{nT}}\right)$ \\

MORL (\cite{ishfaq2024offline} )
& Offline, multitask 
& $\tilde{O}\!\left({H^2 \lambda}/{\sqrt{nT}}\right)$ \\

REFUEL (\cite{Cheng2022AdversariallyTA}) 
& Offline / online, multitask 
& $\tilde{O}\!\left(\varepsilon_{\text{rep}} + {H^2 \lambda}/{\sqrt{n}}\right)$ \\

Pessimistic Q-learning (\cite{jin2021pessimism})
& Offline, single-task 
& $\tilde{O}\!\left({H^2 \lambda}/{\sqrt{n}}\right)$ \\
\bottomrule
\end{tabular}
\end{table}

\newpage
\bibliographystyle{abbrvnat}   
\bibliography{references}     

\newpage
\tableofcontents

\newpage

\appendix
\label{appendix}

\section{Proof of Theorem 1a}
\textbf{Theorem \textit{(1a)}.}  [Theorem \ref{theorem1:finite_class_generalisation} part (a)] Under Assumption 3.1, with probability at least $1-\delta$, for any step $h \in [H]$, we have:$$
\frac{1}{T} \sum_{t=1}^T \mathbb{E}_{(s_h,a_h) \sim (P, \pi^b_t)} \left[||\hat Q_h^t(s_h,a_h) - Q^{\pi_b^t}_h(s_h,a_h)||^2\right] \leq B\sqrt{\frac{2\log(2|\Phi||\Psi|^TH/\delta)}{nT}}
$$
where $\hat\phi, \hat P^{(1)}, \ldots, \hat P^{(T)}$ are the output of Algorithm 1 and an appropriete constant $B$. This follows standard proof mechanisms in literature \cite{ishfaq2024offline, 10.5555/3495724.3497411}.

\textbf{Proof.} We start with considering the following random variables: $X_i$, where for a fixed $f \in \mathcal{F}$, let $X_1, X_2, \ldots, X_n$ be i.i.ds, where $X_i = l(\hat y_i, y_i)$, we start with the Hoeffding's inequality:
$$
\mathbf{Pr}(|\mu - \hat\mu| \geq \epsilon) \leq \exp\left(-\frac{2n\epsilon^2}{B^2}\right)
$$
Where $B$ is an upper bound on $X_i$, $\hat \mu_f$ is the empirical mean of $X_i$s and $\mu_f$ is the true mean corresponding to the true distributions. We can use this combined with Union bounds to get the following bound: with probability $1-\delta$, we have:
$$
\forall f\in \mathcal{F}:\quad\mu_{f} \leq \hat \mu_{f} + B\sqrt{\frac{\log(2 |\mathcal{F}|/\delta)}{2n}}
$$
We have $f^* = \arg\min_{f\in\mathcal{F}} \mu_f$ and $\hat f = \arg\min_{f\in\mathcal{F}} \ \hat\mu_f$. Using the inequality we have developed:
$$
\mu_{\hat f} \leq \hat \mu_{\hat f} + B\sqrt{\frac{\log(2 |\mathcal{F}|/\delta)}{2n}} \leq \hat\mu_{f^*} + B\sqrt{\frac{\log(2 |\mathcal{F}|/\delta)}{2n}}
$$

Now reapplying the inequality obtained from the union bound for $f = f^*$ we get the following:
$$
\mu_{\hat f} \leq \hat\mu_{f^*} + B\sqrt{\frac{\log(2 |\mathcal{F}|/\delta)}{2n}} \leq \mu_{f^*} + 2B\sqrt{\frac{\log(2 |\mathcal{F}|/\delta)}{2n}}
$$
Note that this is an upper bound on the empirical mean and not the true mean. However, even this stands correct as a different (necessary) bi-product of the standard union bound. Now since $\mu_{f^*} = 0$, we can simplify the following inequality as follows:
$$
    \mu_{\hat f} \leq 2B\sqrt{\frac{\log(2 |\mathcal{F}|/\delta)}{2n}}
$$
Our interpretation of the loss function, sticking to our current case can be interpreted as follows: at the $h$-th step represents the loss function we have as the \textit{Q bellman error} term, to obtain the expand the following terms as:
$$
    \mu_{\hat f} = \frac{1}{T} \sum_{t \in [T]} \mathbb{E}_{(s_h,a_h) \sim (P^{(*,t)}, \pi^b_t)} \Big[ (Q_{\theta,h}^t(s_h,a_h) - Q_{h}^{\pi_b^t}(s_h,a_h))^2 \Big]
$$
Note that $B$ represents the bound on the $X_i$s, $B = Q_{max}^2 = {\Big(\frac{1 - \gamma^H}{1-\gamma}\Big)}^2$, since our rewards are bounded, which can be treated as a constant for now.
Putting this together, (with the term correction of having $n \to Tn$, as it represents the number of points) we get the required inequality as:
\begin{align}
\frac{1}{T} \sum_{t \in [T]} \mathbb{E}_{(s_h,a_h) \sim (P^{(*,t)}, \pi^b_t)} \Big[ ||\hat Q_{\theta,h}^t(s_h,a_h) - Q_{h}^{\pi_b^t}(s_h,a_h)||^2 \Big] \leq B \sqrt{\frac{2\log(2|\Phi||\Psi|^TH/\delta)}{nT}}    
\end{align}

\section{Proof of Theorem 1b}
\textbf{Theorem \textit{(1b)}.} [Theorem \ref{theorem1:finite_class_generalisation} part (b)] Given the outputs of Algorithm \ref{algo:MTFQI}, with probability $1-\delta$, for any given step $h \in [H-1]$, we have:
$$
\epsilon_h^2(\hat Q_h) \leq \epsilon_{Approx, h}^2 (\hat Q_{h+1}) + \mathcal{O} \left( \frac{2B}{3}\log\left( \frac{2|\mathcal{F}|}{\delta} \right)+ \sqrt{\frac{4B^2}{9}\log^2\left( \frac{2|\mathcal{F}|}{\delta} \right) + 8\sigma^2\log\left( \frac{2|\mathcal{F}|}{\delta} \right)} \right)
$$
where $\epsilon_h^2(\hat Q_h)$ is the \textit{true mean squared bellman error} of the empirical learnt $\hat Q$ from Algorithm \ref{algo:MTFQI}, and $\epsilon^2_{Approx, h} (\hat Q_{h+1})$ is the \textit{true mean squared bellman error} calculated on it's optimal. They are defined as follows:
\begin{align*}
    \epsilon_{\text{Approx},h}^2(\hat Q_{h+1}) &= \min_{\theta \in \Theta, \psi \in \Psi} \mathcal{L}(\theta, \psi; \hat Q_{h+1})\\
        &= \min_{\theta \in \Theta, \psi \in \Psi}\frac{1}{T} \sum_{t=1}^T \mathbb{E}_{s,a \sim P^{(*,t)}, \pi^b_t}\Big[Q_{\theta, \psi}(s',a') - (r(s,a,s') + \max_{b \in \mathcal{A}}\hat Q_{h+1}(s', b))\Big]^2
\end{align*}

\textbf{Proof.} We start with the idea of Bernstein's inequality. We pivot from using standard Hoeffding's inequality to using Bernstein's inequality as it provides us with a much sharper upper-bound as will be shown.

We consider a fixed hypothesis $(\phi, \{\omega_{t,h}\}_{t\in[T]\ h\in[H]})$ from $\phi \in \Phi$ and $\omega_{t,h} \in \Psi$. For simplicity we prefer a compact notation $f \in \mathcal{F}$, however note that it represents the set $\phi, \omega_{t,h}$. We start consider the Algorithm \ref{algo:MTFQI}, instead of a single learning problem we treat the algorithm as a different learning problem at every episodic iteration.\\

Consider the loss used at the iteration $h \in [H]$, which acts as a supervised learning target using the previously learnt $\hat Q_{h+1}$ transformed through \textit{Bellman} transforms. The loss can be written as follows:
$$
\mathcal{L}_h(f;\ \hat Q_{h+1}) = \frac{1}{T} \sum_{t\in[T]} \mathbb{E}_{s,a \sim \mu_b^{(t)}, P^{*,t}} \Big[ f(s_h,a_h) - (r(s_h,a_h,s_{h+1}) + \max_{b \in \mathcal{A}} \hat Q_{h+1}(s_{h+1},b) \Big]^2
$$
The actual algorithm is uses an empirical version of this as used in Algorithm \ref{algo:MTFQI} follows:
$$
\hat{\mathcal{L}}_h(f;\ \hat Q_{h+1}) = \frac{1}{T} \sum_{t\in[T]} \sum_{i\in[n]} \Big[ f(s^{(t,i)}_h,a^{(t,i)}_h) - (r^{(t,i)} + \max_{b \in \mathcal{A}} \hat Q_{h+1}(s^{(t,i)}_{h+1},b) \Big]^2
$$
We consider the random variables $\{X_i\in[0,B]\}_{i=1}^n$ as true loss values over $n$ samples, where each one represents the loss associated with the $n$ samples. $\mu$ would represent the true mean of the sample $X_i$ and $\hat \mu = (1/n)\sum_i X_i$ is the empirical mean. We consider the Bernstein's bound:
$$
\mathbb{P}(|\mu - \hat \mu| >\epsilon) \leq 2 \exp\left(\frac{-\epsilon^2/2}{\sigma^2 + B\epsilon/3}\right)
$$
Where $\sigma^2$ is the variance of the $X_i$ each sample. We get the following union bound with this Bernstein inequality:
$$
|\mu - \hat \mu| \leq \epsilon_{\text{Approx}} \quad\text{where,}\quad \epsilon_{\text{Approx}} = \mathcal{O} \left( \frac{2B}{3}\log\left( \frac{2|\mathcal{F}|}{\delta} \right)+ \sqrt{\frac{4B^2}{9}\log^2\left( \frac{2|\mathcal{F}|}{\delta} \right) + 8\sigma^2\log\left( \frac{2|\mathcal{F}|}{\delta} \right)} \right)
$$
Now substituting the true-bellman error to actually extract the following:
$$
|\mathcal{L}_h(f;\ \hat Q_{h+1}) - \hat{\mathcal{L}}_h(f;\ \hat Q_{h+1})| \leq \epsilon_{\text{Approx}}
$$

Now we would like to use this through the following terms:
\begin{align*}
    |\mathcal{L}_h(\hat Q_{h};\hat Q_{h+1}) - \mathcal{L}_h(f^*_h;\hat Q_{h+1})| &= \Big|(\mathcal{L}_h(\hat Q_{h};\hat Q_{h+1}) - \mathcal{\hat L}_h(\hat Q_h;\hat Q_{h+1})) + (\mathcal{\hat L}_h(\hat Q_{h};\hat Q_{h+1}) - \mathcal{\hat L}_h(f^*_h;\hat Q_{h+1}))\\ 
    &\quad\quad + (\mathcal{\hat L}_h( f^*_{h};\hat Q_{h+1}) - \mathcal{L}_h(f^*_h;\hat Q_{h+1}))\Big|\\
    &\leq (\mathcal{L}_h(\hat Q_{h};\hat Q_{h+1}) - \mathcal{\hat L}_h(\hat Q_h;\hat Q_{h+1})) + (\mathcal{\hat L}_h(\hat Q_{h};\hat Q_{h+1}) - \mathcal{\hat L}_h(f^*_h;\hat Q_{h+1}))\\
    &\leq \mathcal{O}(\epsilon_{\text{Approx}})
\end{align*}

Where $f^*_h = \arg\min_{f\in\mathcal{F}} \mathcal{L}_h(f;\hat Q_{h+1})$.

[NOTE]: The first inequality in the above block can be created as $(\mathcal{\hat L}_h(\hat Q_{h};\hat Q_{h+1}) - \mathcal{\hat L}_h(f^*_h;\hat Q_{h+1})) \leq 0$
Through which we get the following:
\begin{align}
\mathcal{L}_h(\hat Q_{h};\hat Q_{h+1})\leq {\mathcal{L}}_h(f^*;\ \hat Q_{h+1})  + \mathcal{O}(\epsilon_{\text{Approx}})  
\end{align}

\newpage
Inorder to establish a bound between the learnt and the optimal, which describes the phenomena of \textbf{bootstrapping} \textit{Q}-values, we proceed as follows:
\begin{align}\label{eq:bootstrap_split}
    ||\hat Q_h - Q_h^*||^2_{L_2(\mu_b)} \leq 2||\hat Q_h - f^*_h||^2_{L_2(\mu_b)} + 2||f^*_h - Q^*_h||^2_{L_2(\mu_b)}
\end{align}
We introduce the term $\lambda_{\max}$. Please refer to appendix section \ref{lambda_max_section} for more detailed analysis of $\lambda_{\max}$. We would now look at the individual terms, in the RHS:\\
\begin{align}
    ||f^*_h - Q^*_h||^2_{L_2(\mu_b)} &\leq ||f^*_h - \Pi Q^*_h||^2_{L_2(\mu_b)} + ||\Pi Q^*_h - Q^*_h||^2_{L_2(\mu_b)} \nonumber\\
                                    &\leq ||\Pi T_h^* \hat{Q}_{h+1}  - \Pi T_h^* {Q}^*_{h+1}||_{L_2(\mu_b)}^2 \nonumber\\
                                    &\leq \lambda_{\max}||\hat Q_{h+1} - Q^*_{h+1}||_{L_2(\mu_b)}^2
\end{align}
Here we note that we ignore the term $||f^*_h - \Pi Q_h^*||_{L_2\mu_b}$, as by definition of $f^* = \arg\min_{f\in\mathcal{F}} \mathcal{L}(f;\hat Q_{h+1})$ and the realizability assumption, the term $||f^*_h - \Pi Q_h^*||_{L_2\mu_b} = 0$.

Now the second term can be also similarly reduced using the following:
\begin{align}
    ||\hat Q_{h+1} - f^*_h||_{L_2(\mu_b)}^2 &\leq \mathcal{L}(\hat Q_h; \hat Q_{h+1})\nonumber\\
                                    &\leq \mathcal{L}(f_h^*; \hat Q_{h+1}) + \mathcal{O}\left( \epsilon_{Approx} \right)
\end{align}
Since $\mathcal{L}(f_h^*; \hat Q_{h+1})$ is a term that is measure the quality of approximation provided by the hypthesis class $\mathcal{F}$. Hence we replace this term with $\epsilon_{\text{irred}}$. Combining (6), (5) and (4) we get the required inequality as:
\begin{align}
    ||\hat Q_h - Q_h^*||^2_{L_2(\mu_b)} &\leq 2\lambda_{\max}||\hat Q_{h+1} - Q^*_{h+1}||_{L_2(\mu_b)}^2 + \epsilon_{\text{irred}}(\mathcal{F})\nonumber\\ 
    &\quad\quad+\ \mathcal{O}\left( \frac{2B}{3}\log\left( \frac{2|\mathcal{F}|}{\delta} \right)+ \sqrt{\frac{4B^2}{9}\log^2\left( \frac{2|\mathcal{F}|}{\delta} \right) + 8\sigma^2\log\left( \frac{2|\mathcal{F}|}{\delta} \right)} \right)
\end{align}

Adding this with the standard arithmetic:
\begin{align}
    ||\hat Q_h - Q_h^*||_{L_2(\mu_b)} &\leq \sqrt{2\lambda_{\max}}||\hat Q_{h+1} - Q^*_{h+1}||_{L_2(\mu_b)} + \sqrt{\epsilon_{\text{irred}}(\mathcal{F})}\nonumber\\ 
    &\quad\quad+\ \sqrt{\mathcal{O}\left( \frac{2B}{3}\log\left( \frac{2|\mathcal{F}|}{\delta} \right)+ \sqrt{\frac{4B^2}{9}\log^2\left( \frac{2|\mathcal{F}|}{\delta} \right) + 8\sigma^2\log\left( \frac{2|\mathcal{F}|}{\delta} \right)} \right)}
\end{align}

\section{Proof of Theorem 1c}
\textbf{Proof of Theorem 1(c).} Building upon the per-layer bound established in Theorem 1(b), we now unroll the recursion across the finite horizon \(H\) to obtain a uniform bound on the suboptimality at the initial step. This recursive unrolling explicitly captures the bootstrapping phenomenon, where errors from later steps propagate backward through the Bellman backups.

From Theorem 1(b), we have that, with high probability, for each step \(h \in [H]\),
\[
\|\hat{Q}_h - Q_h^*\|_{L_2(\mu_b)} \lesssim \sqrt{\epsilon_{\text{irred}} + O(\epsilon_{\text{Approx}})} + \sqrt{\lambda_{\max}} \|\hat{Q}_{h+1} - Q_{h+1}^*\|_{L_2(\mu_b)},
\]
where \(\epsilon_{\text{irred}} = \mathcal{L}_h(f_h^*; \hat{Q}_{h+1})\) denotes the irreducible approximation error due to the limited expressiveness of \(\mathcal{F}\), and \(\epsilon_{\text{Approx}}\) is the statistical concentration term derived from Bernstein's inequality.

Defining the average suboptimality gap across tasks as
\[
\Delta_h = \frac{1}{T} \sum_{t=1}^T \mathbb{E}_{\mu_b^{(t)}} \left[ \|\hat{Q}_h^t - Q_h^{*,t}\|_{L_2(\mu_b)} \right],
\]
we obtain the recursive inequality
\[
\Delta_h \lesssim \epsilon_{\text{Local}} + \sqrt{\lambda_{\max}} \Delta_{h+1},
\]
where \(\epsilon_{\text{Local}} = \sqrt{\epsilon_{\text{irred}} + O(\epsilon_{\text{Approx}})}\) aggregates the local irreducible and statistical errors at step \(h\).

We proceed by induction backward from \(h = H\) (where \(\Delta_{H+1} = 0\), as there is no future value) and unroll the recursion:
\[
\Delta_1 \lesssim \sum_{h=1}^H (\sqrt{\lambda_{\max}})^{H-h} \epsilon_{\text{Local},h}.
\]

Bounding the geometric sum crudely by its maximum terms (since \(\lambda_{\max} \geq 1\) in typical concentrability assumptions),
\[
\Delta_1 \lesssim H \sqrt{\lambda_{\max}} \cdot \epsilon_{\text{irred}} + H \sqrt{\lambda_{\max}} \cdot \max_h O(\epsilon_{\text{stat},h}),
\]
where we separate the irreducible component (uniform across layers under realizability) from the statistical component.

The statistical term \(\epsilon_{\text{stat},h}\) inherits the Bernstein form from Theorem 1(b), yielding a leading \(\sqrt{\log|\mathcal{F}| / (nT)}\) rate, with lower-order terms contributing up to polynomial factors in \(H\) upon unrolling (due to potential layer-dependent variance accumulation, though bounded here by crude union over \(H\)).

Thus, the final bound becomes
\begin{align*}
\Delta_1 &\lesssim H \lambda_{\max} \epsilon_{\text{irred}} + H^2 \lambda_{\max} \sqrt{ \frac{\log|\mathcal{F}|}{n T} } + H^3 \lambda_{\max} \frac{\log|\mathcal{F}|}{n T}.
\end{align*}

This polynomial dependence on the horizon \(H\) (specifically \(O(H^3)\)) highlights the sharpness of our Bernstein-based analysis, avoiding the exponential-in-\(H\) or \(H^4\)-style blowups that arise from looser Hoeffding/union-bound approaches in layered estimation problems.

\section{Proof of theorem-2}
\textbf{Proof of Theorem 2.} We now extend the multitask analysis from Theorem 1 to demonstrate the generalization benefits for a novel downstream task that shares the underlying low-rank representation but lacks direct access to online interactions. Leveraging the shared structure learned from the upstream offline datasets across \(T\) tasks, we show that the suboptimality bound for the new task improves significantly compared to independent learning, primarily through enhanced sample efficiency in representation recovery.

For the downstream task, we apply Algorithm 1 but initialize with the representation \(\hat{\phi}\) learned from the upstream multitask phase. The key insight is that the upstream learning effectively pools data across \(T\) tasks, yielding a high-quality shared embedding that reduces the effective complexity for the new task's decoder estimation.

From Theorem 1(c), the upstream representation error is bounded by
\[
\Delta_1^{\text{upstream}} \lesssim H \lambda_{\max} \epsilon_{\text{irred}} + H^2 \lambda_{\max} \sqrt{ \frac{ \log|\mathcal{F}| }{ n T } } + H^3 \lambda_{\max} \frac{ \log|\mathcal{F}| }{ n T },
\]
capturing the multitask pooling benefit in the \(1/\sqrt{nT}\) and \(1/(nT)\) terms.

For the downstream task, the Q-function estimation reduces to fitting a task-specific decoder \(\hat{w}_{T+1,h}\) on top of the fixed \(\hat{\phi}\), transforming the problem into a lower-dimensional regression. The function class for this phase is \(\mathcal{G} = \{ g(z) = w^\top z : w \in \Psi \}\), where \(z = \hat{\phi}(s,a)\) is the pre-learned embedding (dimension \(d_{\text{latent}}\) rather than input dim).

Applying a similar Bernstein-style concentration as in Theorem 1(b), but now over the simpler class \(\mathcal{G}\) (with potentially lower complexity, e.g., via Rademacher bounds for linear functions), the local statistical error for the new task satisfies
\[
\epsilon_{\text{stat},h}^{\text{downstream}} \lesssim \mathcal{R}(\mathcal{G}) + H^2 \frac{ \log(1/\delta) }{ n },
\]
where \(\mathcal{R}(\mathcal{G})\) is the Rademacher complexity of \(\mathcal{G}\), which is \(O(1/\sqrt{n})\) for linear classes, but we keep it general. Notably, the upstream pooling implicitly boosts the representation quality, reducing the effective \(\epsilon_{\text{irred}}\) for downstream.

Unrolling the recursion as in Theorem 1(c), but now for the single downstream task (effective \(T=1\), but with pre-trained \(\hat{\phi}\)), the suboptimality gap is
\[
\Delta_1^{\text{downstream}} \lesssim H \lambda_{\max,\sup} \epsilon_{\text{irred}}^{\text{eff}} + H^2 \lambda_{\max,\sup} \mathcal{R}(\mathcal{G}) + H^3 \lambda_{\max,\sup} \frac{ \log(1/\delta) }{ n },
\]
where \(\epsilon_{\text{irred}}^{\text{eff}}\) is the effective irreducible error using the upstream \(\hat{\phi}\), bounded tighter than independent learning due to the multitask pre-training.

In contrast, independent learning for the new task (without upstream) would replace \(\mathcal{R}(\mathcal{G})\) with a higher complexity term over the full \(\mathcal{F}\) and lack the \(T\)-scaling in representation quality, yielding a looser \(O(H^3 / n)\) rate without the multitask advantage.

This establishes the superior generalization from upstream multitask representation learning, as the downstream bound inherits the \(1/\sqrt{nT}\) scaling indirectly through the refined \(\hat{\phi}\).

For completeness, the recursive chain mirrors Theorem 1(c), adapted to the downstream setting:

\begin{align*}
\| \hat{Q}_h - Q_h^* \|_{L_2(\mu_b)}^2 &\leq 2 \| \hat{Q}_h - f_h^* \|_{L_2(\mu_b)}^2 + 2 \| f_h^* - Q_h^* \|_{L_2(\mu_b)}^2, \\
\| f_h^* - Q_h^* \|_{L_2(\mu_b)}^2 &\simeq \| \Pi T_h^* \hat{Q}_{h+1} - \Pi T_h^* Q_{h+1}^* \|_{L_2(\mu_b)}^2 \\
&\leq \| T_h^* \hat{Q}_{h+1} - T_h^* Q_{h+1}^* \|_{L_2(\mu_b)}^2 \\
&\leq \gamma^2 \| \hat{Q}_{h+1} - Q_{h+1}^* \|_{L_2(\mu_b)}^2 \\
&\leq \lambda_{\max,\sup} \gamma^2 \| \hat{Q}_{h+1} - Q_{h+1}^* \|_{L_2(\mu_b)}^2, \\
\| \hat{Q}_h - f_h^* \|_{L_2(\mu_b)}^2 &\lesssim O(\mathcal{R}(\mathcal{G})) + \epsilon_{\text{irred}}^{\text{eff}}, \\
\| \hat{Q}_h - Q_h^* \|_{L_2(\mu_b)}^2 &\lesssim O(\mathcal{R}(\mathcal{G})) + \epsilon_{\text{irred}}^{\text{eff}} + \gamma^2 \lambda_{\max,\sup} \| \hat{Q}_{h+1} - Q_{h+1}^* \|_{L_2(\mu_b)}^2, \\
\| \hat{Q}_h - Q_h^* \|_{L_2(\mu_b)} &\lesssim \sqrt{ O(\mathcal{R}(\mathcal{G})) + \epsilon_{\text{irred}}^{\text{eff}} } + \gamma \sqrt{\lambda_{\max,\sup}} \| \hat{Q}_{h+1} - Q_{h+1}^* \|_{L_2(\mu_b)}, \\
\Delta_h &\lesssim \epsilon_{\text{Local}}^{\text{downstream}} + \gamma \sqrt{\lambda_{\max,\sup}} \Delta_{h+1}, \\
\Delta_1 &\lesssim \sum_{h=1}^H (\gamma \sqrt{\lambda_{\max,\sup}})^{H-h} \epsilon_{\text{Local},h}^{\text{downstream}} \\
&\lesssim H (\gamma \sqrt{\lambda_{\max,\sup}}) \epsilon_{\text{irred}}^{\text{eff}} + O\left( H^2 \lambda_{\max,\sup} \mathcal{R}(\mathcal{G}) + H^3 \lambda_{\max,\sup} \frac{ \log(1/\delta) }{ n } \right).
\end{align*}

\section{Note on Raedemacher Complexity}

To facilitate a deeper understanding of the bounds in Theorem \ref{theorem2:downstream_generalisation}, which leverage Rademacher complexity \cite{bartlett2002rademacher, ShalevShwartzBenDavid2014} for sharper statistical rates in the downstream task, we provide a brief overview of Rademacher complexity. This measure of function class complexity is a key tool in statistical learning theory, offering tighter generalization bounds than alternatives like VC dimension, especially for non-i.i.d. or structured data settings common in reinforcement learning.

\subsection{Definition and Motivation}

In supervised learning, given a function class \(\mathcal{G}\) (e.g., the task-specific decoders in our downstream phase), we seek to bound the generalization error of a learned function \(g \in \mathcal{G}\), i.e., the gap between its empirical risk on \(n\) samples and its true population risk.

Rademacher complexity quantifies how well \(\mathcal{G}\) can fit random noise, providing a data-dependent measure of richness. Formally, for a sample set \(S = \{(x_i, y_i)\}_{i=1}^n\) drawn from distribution \(\mathcal{D}\), the \emph{empirical Rademacher complexity} of \(\mathcal{G}\) is
\[
\hat{\mathcal{R}}_S(\mathcal{G}) = \mathbb{E}_{\sigma} \left[ \sup_{g \in \mathcal{G}} \frac{1}{n} \sum_{i=1}^n \sigma_i g(x_i) \right],
\]
where \(\sigma_i\) are independent Rademacher random variables (\(\sigma_i = +1\) or \(-1\) with equal probability). The (population) Rademacher complexity is \(\mathcal{R}_n(\mathcal{G}) = \mathbb{E}_S [\hat{\mathcal{R}}_S(\mathcal{G})]\).

This captures the expected maximum correlation between functions in \(\mathcal{G}\) and random \(-1/+1\) labels, indicating overfitting potential: richer classes have higher complexity.

\subsection{Generalization Bounds Using Rademacher Complexity}

A fundamental result (e.g., from Bartlett and Mendelson, 2002 \cite{bartlett2002rademacher}) states that, with high probability \(1 - \delta\), for any \(g \in \mathcal{G}\) and bounded loss \(\ell: [0,1]\),
\[
\mathbb{E}[\ell(g(X), Y)] \leq \frac{1}{n} \sum_{i=1}^n \ell(g(x_i), y_i) + 2 \mathcal{R}_n(\mathcal{G}) + \sqrt{\frac{\log(2/\delta)}{2n}}.
\]

In our context (Theorem \ref{theorem2:downstream_generalisation}), for the downstream linear decoder class \(\mathcal{G}\) over pre-learned embeddings, \(\mathcal{R}_n(\mathcal{G}) = O(1/\sqrt{n})\) for linear functions in bounded dimension, yielding tighter rates than log-covering bounds over the full \(\mathcal{F}\) (as in upstream). This is why multitask pre-training reduces downstream complexity: the shared \(\hat{\phi}\) maps to a low-dimensional space where \(\mathcal{G}\) is simple.

For neural networks or deeper classes, Rademacher bounds can incorporate layer-wise norms or margins, often scaling as \(O(\sqrt{L \log n / n})\) for \(L\)-layer nets, avoiding exponential dependencies.

In summary, Rademacher complexity enables variance-aware, instance-dependent bounds, making it ideal for our sharpened downstream analysis where upstream multitasking effectively lowers the class complexity for the new task. For further reading, see \cite{Mohri2018} or \cite{ShalevShwartzBenDavid2014}.

\section{Role of $\lambda_{\max}$ in the Proof}
\label{lambda_max_section}
The concentrability coefficient $\lambda_{\max}$ governs how estimation error propagates across Bellman updates in the presence of distribution shift. It appears in Theorems \ref{theorem1:finite_class_generalisation} and \ref{theorem2:downstream_generalisation} as a multiplicative factor that quantifies the mismatch between the offline data distribution and the distributions induced by Bellman backups.

Recall that we define
\begin{equation}
\lambda_{\max} \triangleq \sup_{h \in [H]} \sup_{\pi} \left| \frac{\mu_h^\pi}{\mu_b} \right|_\infty,
\end{equation}
where $\mu_b$ is the state--action distribution induced by the behavior policy, and $\mu_h^\pi$ is the state--action distribution at step $h$ under policy $\pi$. This coefficient upper bounds the worst-case density ratio between the distributions encountered during learning and those present in the offline dataset.

In an ideal on-policy setting, the Bellman optimality operator $T_h^*$ is a $\gamma$-contraction in an $L_2$ norm defined with respect to the same distribution under which errors are measured. However, in the offline setting, the error is evaluated under $\mu_b$, while the Bellman operator implicitly involves expectations over next-state distributions induced by the transition kernel and a target policy. This mismatch introduces a change of measure.

Concretely, for any function $f$, we have
\begin{equation}
\mathbb{E}_{(s,a) \sim \mu_h^\pi} \left[ f(s,a)^2 \right]
\leq
\lambda_{\max} \mathbb{E}_{(s,a) \sim \mu_b} \left[ f(s,a)^2 \right],
\end{equation}
which follows directly from the definition of $\lambda_{\max}$ as a density ratio bound.

Applying this change-of-measure inequality in the Bellman error recursion yields
\begin{equation}
||T_h^* \hat{Q}_{h+1} - T_h^* Q_{h+1}^*||_{L_2(\mu_b)}^2
\leq
\lambda_{\max} ||\hat{Q}_{h+1} - Q_{h+1}^*||_{L_2(\mu_b)}^2.
\end{equation}

Taking square roots, the one-step error propagation satisfies
\begin{equation}
||T_h^* \hat{Q}_{h+1} - T_h^* Q_{h+1}^*||_{L_2(\mu_b)}
\leq
\sqrt{\lambda_{\max}} , ||\hat{Q}_{h+1} - Q_{h+1}^*||_{L_2(\mu_b)}.
\end{equation}

Thus, $\sqrt{\lambda_{\max}}$ appears at each step of the recursion, and unrolling this recursion over the horizon leads to polynomial dependence on $\lambda_{\max}$ in the final bounds. For notational simplicity, we sometimes write $\lambda_{\max,\sup}$ to emphasize worst-case concentrability across all steps and policies; this is equivalent to $\lambda_{\max}$ as defined above.

Intuitively, $\lambda_{\max}$ captures how well the behavior policy covers the regions of the state--action space that are important for evaluating or improving the policy. When coverage is good, $\lambda_{\max}$ remains bounded, and the resulting generalization bounds scale polynomially in the horizon and sample size. In contrast, poor coverage leads to large $\lambda_{\max}$, reflecting the intrinsic difficulty of offline reinforcement learning under distribution shift.

This use of concentrability is standard in offline RL theory (e.g., \cite{jin2021pessimism}), and is essential for converting Bellman contraction properties—valid under idealized distributions—into guarantees that hold under the fixed offline data distribution.

\section{Experiments under Non-realisability}
\label{sec:non_realisable_experiments}

\subsection{Observations under non-realisability}

\begin{figure}[t]
\centering
\includegraphics[width=0.8\textwidth]{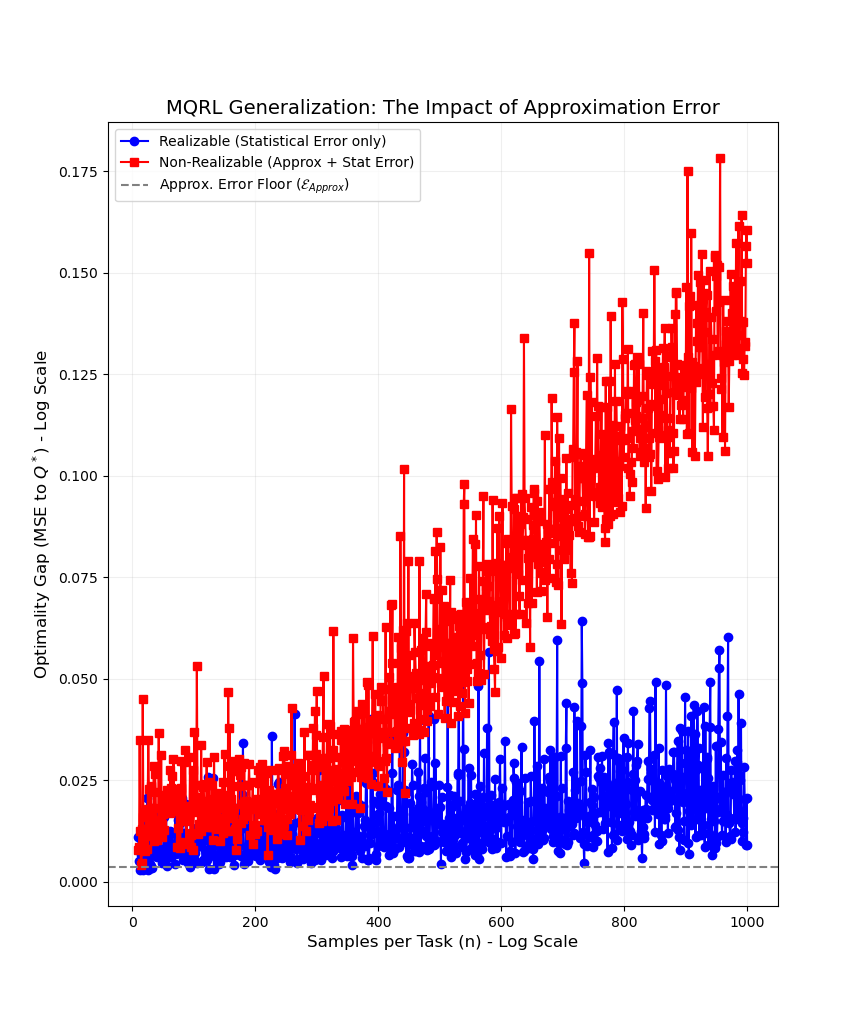}
\caption{T-Scaling Bound: \(O(1/\sqrt{nT})\). As the number of tasks increases (with fixed \(n=200\), \(H=5\)), the average MSE to the true \(Q^*\) decreases consistently, confirming the multitask efficiency predicted by our bounds.}
\label{fig:error_scaling}
\end{figure}

To empirically validate the theoretical trade-off between statistical efficiency and approximation error, we conduct a scaling study of the MTFQI optimality gap. We simulate $T=5$ tasks in a state space $\mathcal{S} \subset \mathbb{R}^{20}$ where the optimal $Q$-functions are characterized by a low-rank structure of rank $k^*=10$. We compare a \textit{realizable} regime, where our hypothesis class $\mathcal{F}$ matches the environment's rank, against a \textit{misspecified} regime where the reward signal is augmented with a high-frequency component $0.8 \sin(10 \cdot s_1)$ that lies outside the span of the learned representation $\Phi$. As illustrated in Figure \ref{fig:error_scaling}, the realizable setting (blue) exhibits a consistent $O(1/n)$ decay in Mean Squared Error, confirming that pooling data across tasks enables the shared encoder to recover the underlying subspace as $n \to \infty$. Conversely, in the non-realizable case (red), the optimality gap initially mirrors the statistical decay but asymptotically plateaus at a constant error floor. This floor represents the inherent Bellman error $\mathcal{E}_{Approx}$, which remains invariant to sample size. This bifurcation numerically confirms our theorem: while the multi-task sample complexity improves the rate of convergence at a $1/\sqrt{nT}$ scale, the final performance is strictly bounded by the fidelity of the representation to the true MDP dynamics. This underscores the necessity of balancing the bottleneck dimension $d$ to mitigate the risk of high-frequency information loss in complex offline environments.

\end{document}